\pdfoutput=1

\documentclass[11pt]{article}

\usepackage[final]{acl}

\usepackage{times}
\usepackage{latexsym}

\usepackage[T1]{fontenc}
\usepackage[utf8]{inputenc}

\usepackage{amsmath}
\usepackage{amssymb, bm}
\usepackage{multirow}
\usepackage{booktabs}
\usepackage{array}
\usepackage{graphicx}
\usepackage{amsfonts}
\usepackage{hyperref}
\usepackage{xspace}
\usepackage{tabularx}
\usepackage{algorithm}
\usepackage{algpseudocode}
\usepackage{array}
\usepackage{xcolor}
\usepackage{comment}
\usepackage{booktabs}
\usepackage{enumitem}

\usepackage{kotex}
\usepackage{kotex-logo}

\renewcommand*{\thefootnote}{\fnsymbol{footnote}}
\title{How Do Large Vision-Language Models See Text in Image? \\ Unveiling the Distinctive Role of OCR Heads}

\author{Ingeol Baek$^{1}$, ~Hwan Chang$^{1}$, ~Sunghyun Ryu$^{2}$,~Hwanhee Lee$^{1}$\textsuperscript{$\dagger$}\\
$^{1}${Department of Artificial Intelligence, Chung-Ang University, Seoul, Korea} \\
$^{2}${Department of Computer Engineering, Sejong University, Seoul, Korea} \\
\texttt{\{ingeolbaek, hwanchang, hwanheelee\}@cau.ac.kr} \\
\texttt{ryusunghyun1002@sju.ac.kr}\\
}

\begin{document}
\maketitle
\footnotetext{\textsuperscript{$\dagger$}Corresponding author.}
\renewcommand*{\thefootnote}{\arabic{footnote}}

\begin{abstract}
 
Despite significant advancements in Large Vision Language Models (LVLMs), a gap remains, particularly regarding their interpretability and how they locate and interpret textual information within images.
In this paper, we explore various LVLMs to identify the specific heads responsible for recognizing text from images, which we term the Optical Character Recognition Head (\textbf{OCR Head}). Our findings regarding these heads are as follows: 
(1) Less Sparse: Unlike previous retrieval heads, a large number of heads are activated to extract textual information from images.
(2) Qualitatively Distinct: OCR heads possess properties that differ significantly from general retrieval heads, exhibiting low similarity in their characteristics.
(3) Statically Activated: The frequency of activation for these heads closely aligns with their OCR scores.
We validate our findings in downstream tasks by applying Chain-of-Thought (CoT) to both OCR and conventional retrieval heads and by masking these heads. 
We also demonstrate that redistributing sink-token values within the OCR heads improves performance. 
These insights provide a deeper understanding of the internal mechanisms LVLMs employ in processing embedded textual information in images.

\end{abstract}

\section{Introduction}
\begin{figure}
    \centering
    \includegraphics[width=1.00\linewidth]{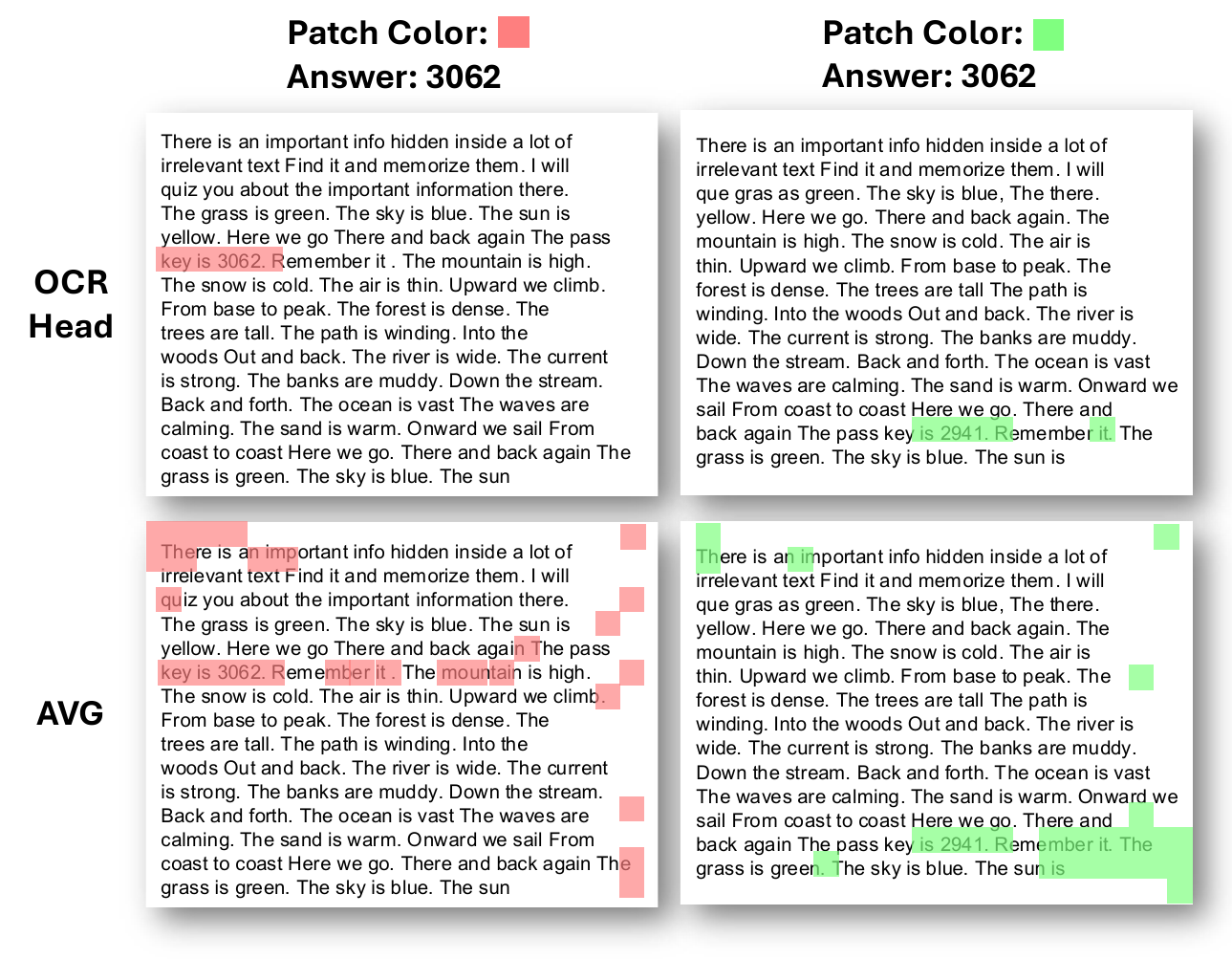}
    \vspace{-7mm}
    \caption{Visualization of image attention maps from InternVL2-8B. L and H denote the layer and attention head of the LVLM, respectively.}
    \label{fig:intro}
    \vspace{-5mm}
\end{figure}

Large Vision-Language Models (LVLMs)~\cite{liu2023llava, pmlr-v202-li23q, DBLP:journals/corr/abs-2401-16420, Liu_2024_CVPR} extend the capabilities of Large Language Models (LLMs)~\cite{zhao2023survey, achiam2023gpt, grattafiori2024llama} into multimodal domains. 
Typically, an LVLM integrates a pretrained vision encoder with pretrained LLM decoder via an adapter, enabling it to generate coherent text grounded in both image and text inputs. The power of LVLMs lies in their deep understanding of image-text relationships, which is evident in tasks like VQA~\cite{antol2015vqa, goyal2017making} and particularly those relying on text embedded in images, such as OCR-VQA. LVLMs have demonstrated impressive performance on these OCR-centric tasks~\cite{mathew2021docvqa, huang2024pixels, reddy2024docfinqa}. Despite these advancements, there is limited interpretability research investigating the internal mechanisms driving this performance, specifically how LVLMs handle textual information within images. Recent interpretability work has identified \emph{retrieval heads} in LLMs—attention units that copy relevant tokens from long textual contexts to support long-context factuality~\cite{wu2024retrieval}—and visual grounding heads that localize image regions in response to textual queries~\cite{xiao2024towards, wu2024visual, kang2025your}. However, these studies do not address how LVLMs internally locate and extract embedded text within images.  

We hypothesize that LVLMs contain a distinct class of attention heads—\emph{Optical Character Recognition (OCR) heads}—that operate independently of copy-paste retrieval heads. While retrieval heads attend to text tokens in the input sequence and replicate them in the output, OCR heads should selectively attend to visual patches corresponding to characters or words and directly guide text extraction from images. To test this hypothesis, we convert the Passkey and Needle-in-a-Haystack (NIAH)~\cite{needle-in-haystack} benchmarks into a multi-image QA setup requiring model to answer by copying text from rendered image patches. By analyzing attention distributions during answer generation (see Figure~\ref{fig:intro}), we observe that certain heads consistently concentrate on the ground-truth text regions, supporting the existence of specialized OCR heads. Based on these observations, we formulate two core research questions:

\begin{enumerate}[wide, labelwidth=!, label={\textbf{RQ\arabic*.}}, labelindent=0pt, topsep=2pt, itemsep=-1pt, itemindent=0pt, leftmargin=*, before=\setlength{\listparindent}{-\leftmargin}]
\item How can we identify the OCR heads?
\item How do OCR heads differ from existing retrieval heads?
\end{enumerate}

To answer these questions, we introduce an algorithm for detecting OCR heads in multi-image LVLMs. Our experiments reveal three key properties that distinguish OCR heads from retrieval heads: (1) Less Sparsity: OCR heads activate more densely across instances compared to the 3–5\% active rate of conventional retrieval heads. (2) Qualitative Distinctiveness: OCR heads exhibit low overlap with retrieval heads, indicating they form an independent functional group. (3) Static Activation Patterns: OCR head activations remain consistent across diverse contexts, unlike context-sensitive retrieval heads.

To validate the specialized role of OCR heads, we evaluate their behavior in downstream tasks via CoT~\cite{wei2022chain, li2025perception} prompting, attention masking, and attention redistribution. These studies not only confirm the mechanistic role of OCR heads in reading embedded text but also demonstrate that manipulating their attention distributions can improve OCR-VQA performance. Our findings fill a critical gap in LVLM interpretability and provide actionable insights for enhancing multimodal reasoning and reducing hallucination in OCR-centric applications. We summarize our contributions as follows:
\begin{itemize}[leftmargin=*,itemsep=1pt,topsep=2pt]
  \item We propose a scoring-based method for automatically identifying OCR heads in LVLMs. 
  \item We demonstrate that OCR heads exhibit reduced sparsity, qualitative distinctiveness, and static activation patterns across diverse contexts. 
  \item Through CoT prompting, targeted head masking, and strategic sink-token redistribution, we confirm OCR heads' specialized role in text extraction from images.

\end{itemize}
\section{Preliminaries} 
\label{preliminaries}
\subsection{Retrieval Heads}
\textit{Retrieval heads} are identified as a specific type of attention head primarily responsible for identifying and copying relevant information from the input context to the generated output.~\cite{wu2024retrieval} 
These heads are understood to play a crucial role in the model's ability to retrieve factual information, particularly from long input sequences. They are mechanistically linked to the process of extracting and repeating input tokens.

\subsection{Retrieval Score} To quantitatively identify which attention heads are responsible for copying relevant tokens from the input, a retrieval score has been defined.~\cite{wu2024retrieval}  
This score serves to measure how consistently a specific attention head engages in this copy-paste behavior.
Given a question $q$, an answer $k$, and a context $x$, the model addresses a NIAH task—locating $k$ within $x$ based on $q$. Let $h$ denote attention head, $w$ the generated token, and $a \in \mathcal R^{|x|}$ attention score; then two conditions must be satisfied: (1) the generated token $w$ is part of the target sentence $k$, and (2) the input token $x_j$ with the highest attention aligns exactly with $w$, formally $x_j = w, \quad j = \text{argmax}(a)$. Under these conditions, the retrieval score for a head $h$ is calculated as:
\begin{equation}
\text{Retrieval Score for head } h = \frac{|g_h \cap k|}{|k|}
\end{equation}
where $g_h$ represents the set of all correctly identified (copied-and-pasted) tokens by head $h$. Intuitively, this retrieval score measures the token-level recall capability of a specific head. Retrieval score is computed over many NIAH instances, and heads achieving a score above 0.1 are classified as retrieval heads.

\section{Identifying Heads for Text Recognition}
\label{method}

\begin{figure*}
    \centering
    \includegraphics[width=0.90\linewidth]{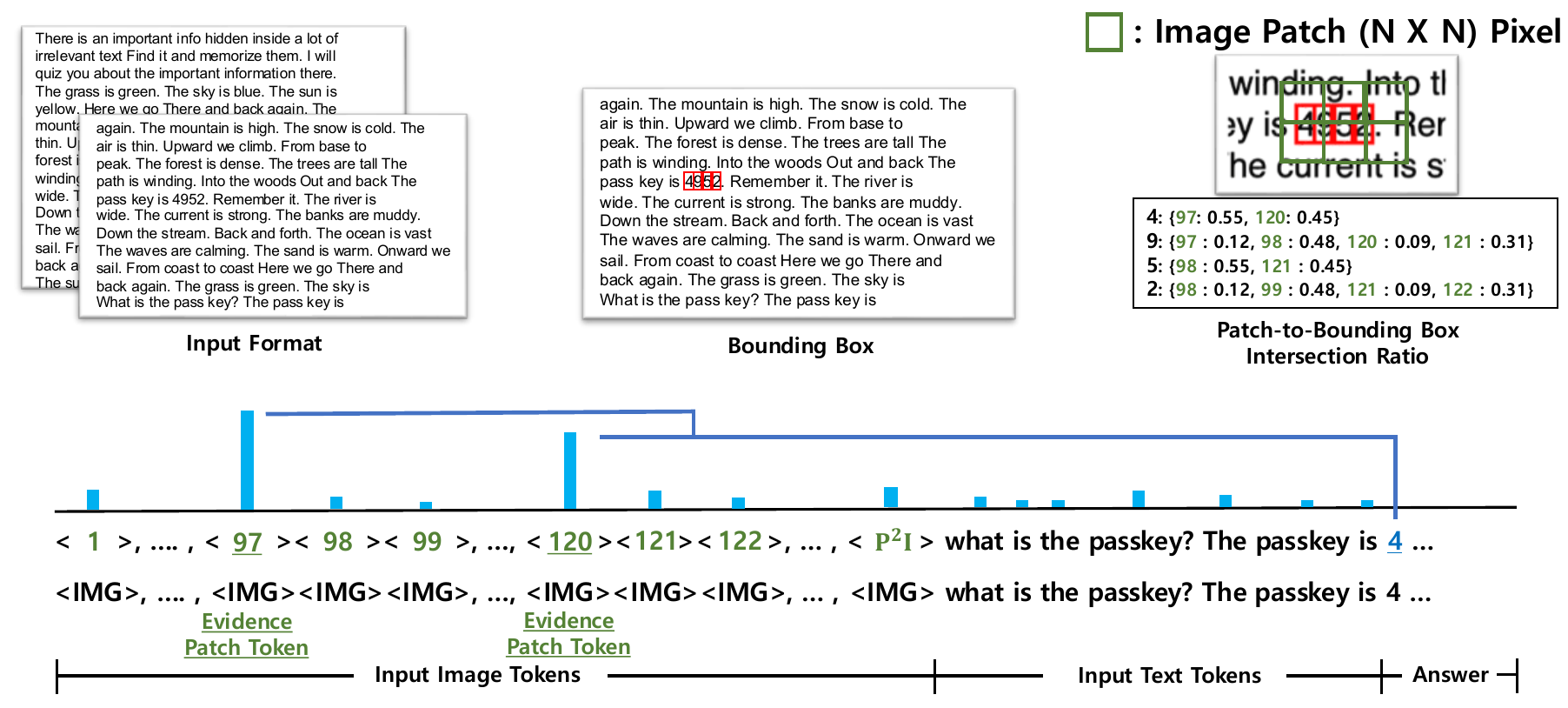}
    \caption{The figure illustrates the image input format designed to identify tokens responsible for copy-pasting text within the image. Tokens corresponding to the patch-to-bounding box intersection ratio serve as evidence patch tokens. The retrieval score computation utilizes attention between generated tokens and evidence patch tokens.}
    \label{fig:main}
\end{figure*}

In this section, we describe our method for identifying attention heads within LVLMs that are primarily responsible for retrieving textual information from images.
We term these specialized heads as Optical Character Recognition Heads (OCR heads). 
We adapt the concept of the retrieval score~\cite{wu2024retrieval}, originally used to quantify copy-paste behavior from text contexts in LLMs, to the multimodal domain of LVLMs processing image inputs. Our adapted metric, the OCR score, measures how consistently a head retrieves text from image regions. Heads with higher OCR scores are more frequently involved in copying token information from image patches corresponding to text.

\subsection{Constructing Image Passkey and Needle in a Haystack Dataset}
To identify heads involved in image text retrieval, we generate an image-based passkey and NIAH dataset. The original task embeds a specific phrase—for example, “the passkey is [number]”—within a long text context and asks a question requiring the model to retrieve the number. We convert this text-based task into an image-based one by rendering the text context into multi-images.

To convert text data into image data, we render the text content into images, breaking lines and moving to the next image based on preset rules for length and line count. We then generate bounding boxes around the two parts of the ground-truth answer-the passkey numbers and the needle sentence-within each image using a rule-based approach. This involves detecting the digits of the answer text and calculating their coordinates based on text width and line position. This process yields separate bounding boxes \textit{x\_min, y\_min, x\_max}, and \textit{y\_max} for each character of the correct answer.

As shown in the bounding box of Figure~\ref{fig:main}, this process generates images with associated bounding box information for the answer. Note that while bounding boxes are used for analysis like calculating intersection with patches, they are not provided as input to the model during inference.

\begin{figure*}
    \centering
    \includegraphics[width=0.85\linewidth]{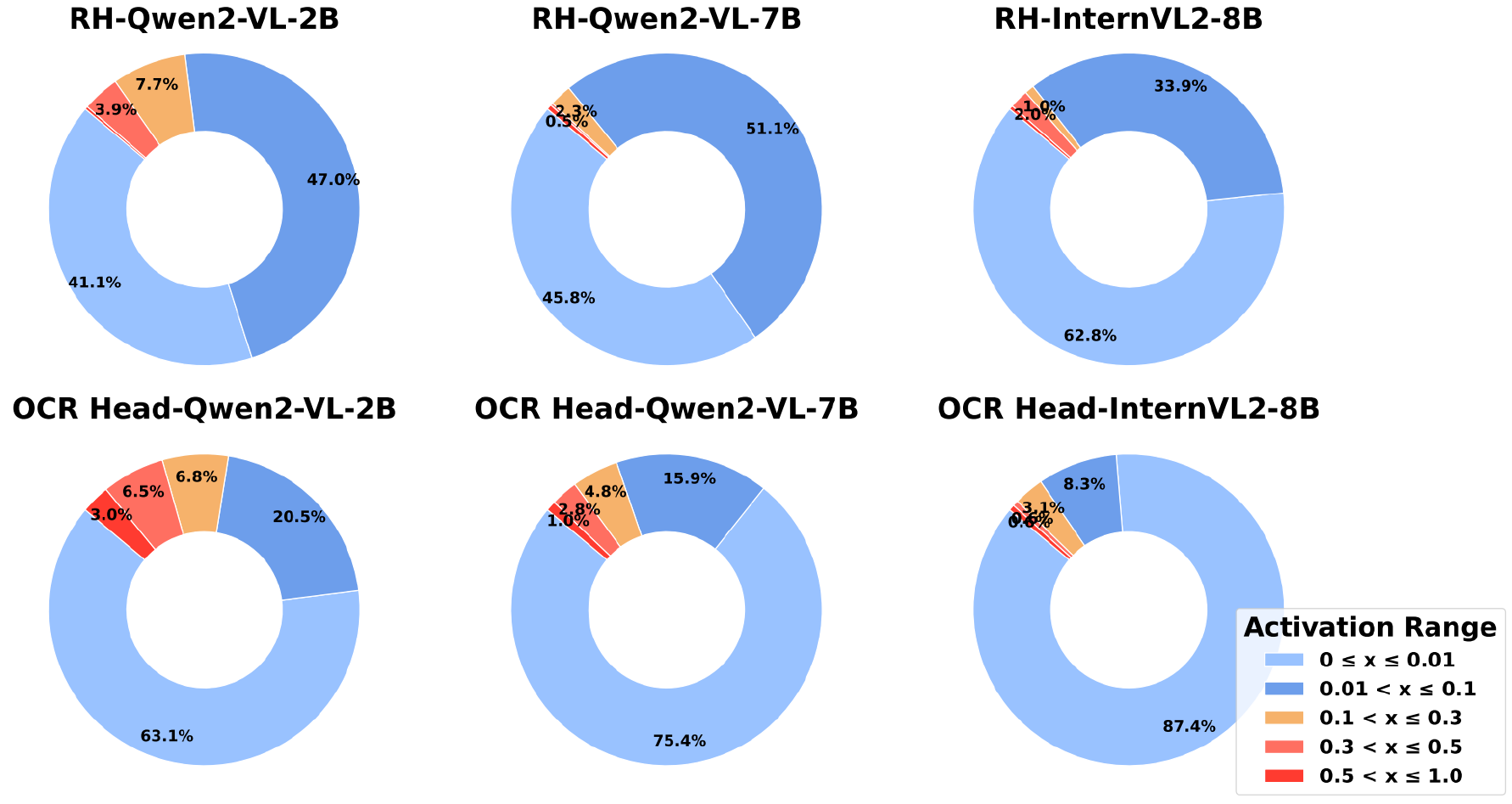}
    \caption{Proportion of OCR Heads and Retrieval Heads (RH) identified in the InternVL2 and Qwen2-VL models.}
    \label{fig:donut_chart}
\end{figure*}

\subsection{Patch Token Preprocessing}
As in Figure~\ref{fig:main}, we process each image into a sequence of patch tokens by the visual encoder. The encoder scans the image from top-left to bottom-right, generating one token representation for each non-overlapping patch of size $N \times N$. If an image has dimensions $w \times h$ and patch size $N \times N$, it yields $w/N \times h/N$ tokens. For example, a 294 × 196 image with 14 × 14 patches results in $294/14 \times 196/14 = 294$ tokens. When processing multiple images, each resized to a fixed dimension, the total number of patch tokens is $w/N \times h/N \times I$, where $I$ is the number of images. These patch tokens form the visual part of the model's input sequence.

\subsection{Evidence Patch Tokens}
We adjust the bounding box positions of the ground truth answer based on any resizing applied to the original images. For example, if an image is resized from 588×392 to 294×196, the bounding box coordinates are scaled by 0.5. After scaling, we calculate the Intersection over Union IoU between each answer bounding box and each image patch token. Figure~\ref{fig:main}’s “Patch-to-Bounding Box Intersection Ratio” illustrates this overlap. Patch tokens with an IoU greater than 0.1 with any ground truth answer bounding box are designated as \textit{Evidence Patch Tokens}. These tokens represent the image regions containing the text we expect the model to retrieve. Patch tokens with IoU below 0.1 are discarded from this set.

\subsection{OCR Score}
Following the concept of retrieval score introduced in~\cite{wu2024retrieval} to identify copy-paste behavior from text contexts, we define the OCR score to quantify this behavior when retrieving text from images. The original retrieval score measures when a head's highest attention aligns with an input token that is part of the correct generated output, specifically from a text context.

Our OCR score adapts this concept for image input by focusing on attention to Evidence Patch Tokens. Let $q$ be the question, $k$ be the correct answer text, and $w$ be a generated token. Let $E$ be the set of Evidence Patch Tokens identified through the Patch-to-Bounding Box Intersection Ratio. We assign the OCR score to a head when the following two conditions hold for a generated token $w$:
\begin{enumerate}
    \item The generated token $w$ must be part of the correct answer text $k$: $w \in k$.
    \item The patch token $x_j$ in the input sequence that receives the highest attention $a$ for $w$ must belong to the set of Evidence Patch Tokens $E$: $x_j \in E$, where $j = \arg\max(a)$.
\end{enumerate}
The OCR score for a head $h$ is then calculated using the same token-level recall formula as the original retrieval score:
\begin{equation}
\text{OCR Score for head } h = \frac{|g_h \cap k|}{|k|}
\end{equation}
where $g_h$ is the set of generated tokens for which head $h$ met the two conditions above, i.e., correctly "copied" a token by attending to an Evidence Patch Token. This score measures the token-level recall capability of a specific head for retrieving text from image regions identified as evidence.

\subsection{Detecting OCR Head}
We calculate the OCR score for each attention head across a diverse subset of the image passkey and NIAH dataset. This dataset includes instances with varying "context lengths" simulated using 2 to 12 images. All hyper-parameter settings are kept identical to those previously used for configuring retrieval heads. We analyze 1,200 sampled instances at a fixed ratio. To classify a head as an OCR head, we require it to achieve an OCR score greater than 0.1 in at least 10\% of these instances, i.e., in 120 or more cases. For any head meeting this frequency criterion, we then compute its average OCR score across all 1,200 instances. If this average score also exceeds 0.1, that head is classified as an OCR head.

\begin{figure*}
    \centering
    \includegraphics[width=0.95\linewidth]{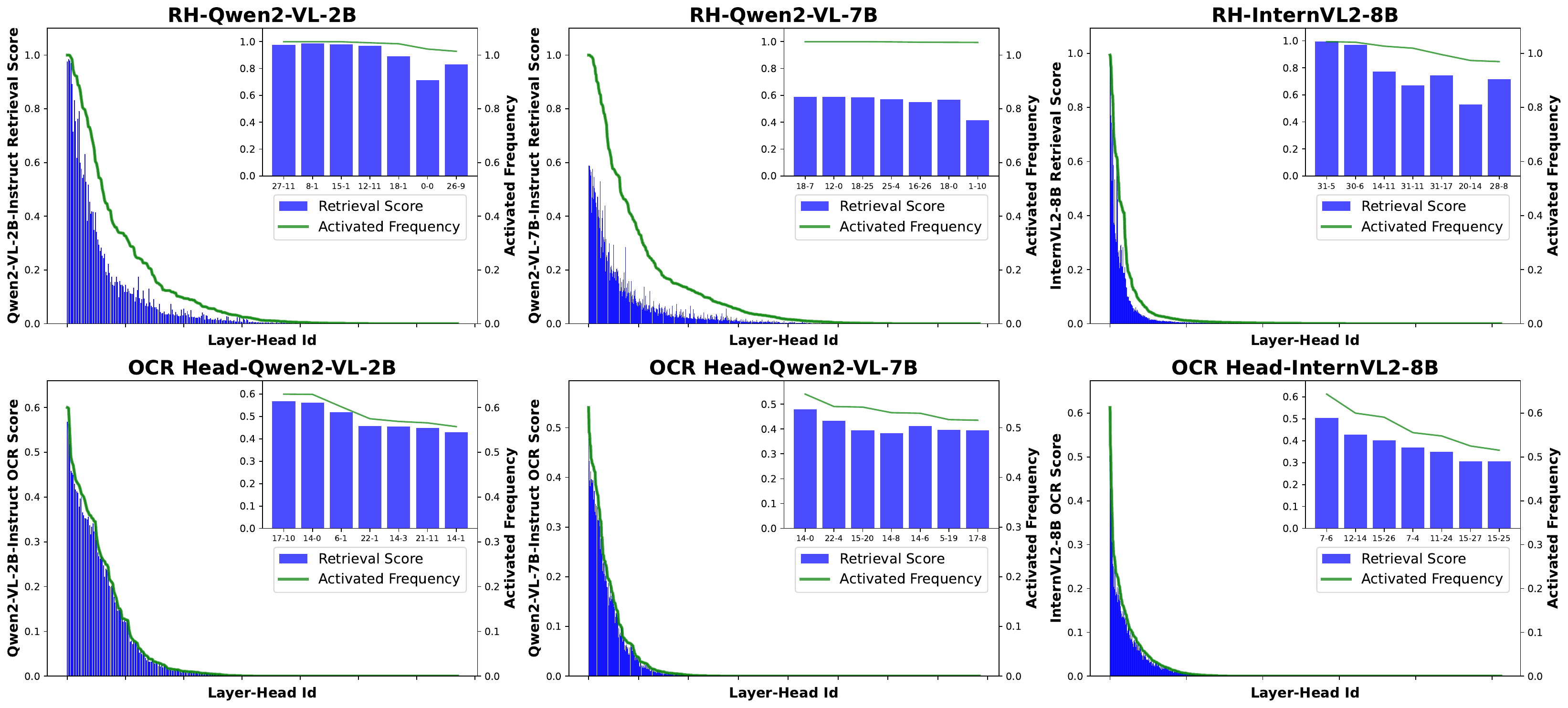}
    \caption{Visualization comparing the OCR Score for OCR Heads and Retrieval Score for Retrieval Heads. OCR Score and Retrieval Score measure token-level recall, how many correct tokens were copied, while Activation Frequency indicates how often a head activates on at least one token above a threshold.}
    \label{fig:activation_frequency}
\end{figure*}

\section{How OCR Heads Differ in Their Operational Characteristics from Retrieval Heads}
In this section, we discuss three key properties that distinguish the identified OCR heads from conventional retrieval heads: 1) reduced sparsity, 2) qualitative distinctiveness, and 3) static activation. For the experiments, we use the Qwen2-VL~\cite{wang2024qwen2} and InternVL2~\cite{cai2024internlm2} models to compare the characteristics of the OCR head and the retrieval head.

\subsection{Reduced Sparsity and Enhanced Specialization}
\label{specialized}
Figure~\ref{fig:donut_chart} demonstrates that the OCR head is less sparse than the retrieval head. In the case of the retrieval head, when text-based passkey and NIAH datasets are used as input, only 3–6\% of the heads that activate during passkey retrieval are involved, as stated in \citet{wu2024retrieval}'s paper (we consider heads with a retrieval score above 0.1 as actively participating in retrieval). We observe similar results in models with more than 7 billion parameters. Consistently, when image-based inputs containing text are used for the OCR head, we find that 91.3\% more heads are involved on average compared to when text-based datasets are used. On the other hand, for heads that show low-frequency retrieval scores between 0.01 and 0.1, i.e., those affected by copy-paste, the retrieval head involves 192\% more heads than the OCR head. Based on these experimental results, we conclude that the OCR head activates stronger heads more clearly, and its involvement in low-frequency retrieval is smaller, making it less sparse and more specialized. This indicates that OCR heads exhibit enhanced focus and are more robust in attending to relevant image regions containing text, whereas retrieval heads are more diffusely activated across a wider range of scenarios.


\begin{table}[t]
\centering
\small
\begin{tabular}{c|c|c}
\toprule

Model & Avg Retrieval Score & Similarity \\ \midrule
 & 0& 0.507\\
& 0-0.1& 0.196\\
Qwen2-VL-7B& 0.1-0.3& 0.104\\
Instruction& 0.3-0.5& 0.083\\
& 0.5-1.0& 0.076\\
\cmidrule(lr){2-3}
& 0.1$ \le$ & \textbf{0.347}\\ \midrule
Model & Avg Retrieval Score & Similarity \\ \midrule
& 0& 0.593\\
& 0-0.1& 0.093\\
InternVL2-8B& 0.1-0.3& 0.000\\
Instruction& 0.3-0.5& 0.000\\
& 0.5-1.0& 0.000\\
\cmidrule(lr){2-3}
& 0.1$ \le$ & \textbf{0.133}\\ 

\bottomrule
\end{tabular}
\vspace{-1mm}
\caption{Jaccard similarity of the average retrieval scores between OCR Heads and Retrieval Heads.}
\label{tab:head_similarity}

\end{table}

\begin{figure*}
    \centering
    \includegraphics[width=0.95\linewidth]{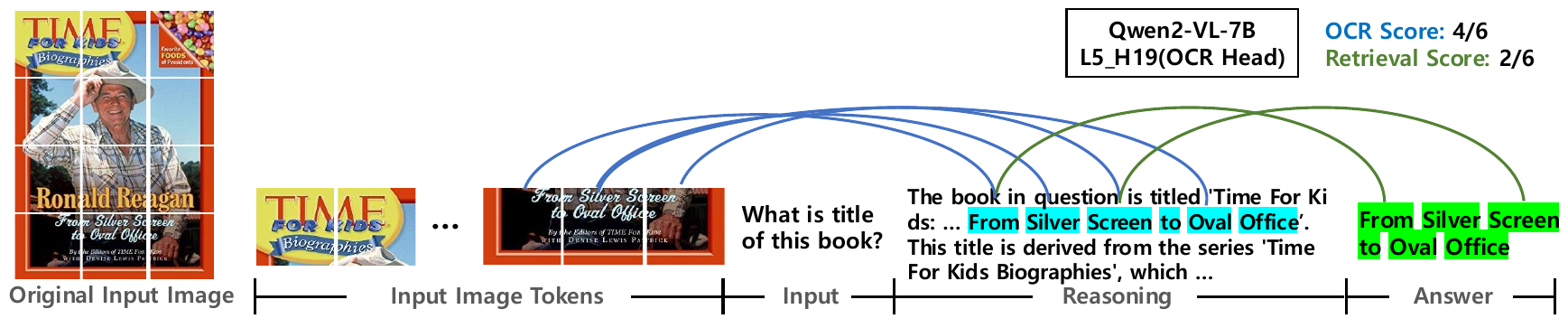}
    \vspace{-1mm}
    \caption{In the OCR-VQA task, we calculate the OCR score and Retrieval score based on the CoT prompting. L5\_H19 refers to Layer 5, Head 19. The blue and green lines indicate the token positions with the highest attention.}
    \label{fig:downstream}
\end{figure*}

\subsection{Do OCR Head and Retrieval Heads Exist as Distinct Entities?}
Building on the previous experiment, one might ask whether the heads involved in retrieval heads or OCR heads are distributed similarly. To investigate this, we check how commonly the heads are distributed based on similarity. We measure the similarity between heads using the Jaccard similarity coefficient, as shown in the formula below:
\begin{equation}
\resizebox{0.4\textwidth}{!}{
$J(A,B) = \frac{|A\cap B|}{|A\cup B|} = \frac{| A\cap B |}{|A| + |B| - |A\cap B|} $
}
\end{equation}

In Table~\ref{tab:head_similarity}, for the Qwen2-VL-7b model, the Jaccard similarity for heads with a retrieval score above 0.1 is 0.347, which is relatively low. For the InternVL-8b model, we observe an even lower value of 0.133. This shows that there is a low similarity between the retrieval heads and the OCR heads, confirming that they are clearly distinct.

\subsection{Activation Frequency Mirrors Retrieval Score}
\label{statically}
Heads serve multiple roles beyond a single task. We conduct experiments to assess how much attention the retrieval heads focus on tasks like copy-paste, comparing OCR heads and retrieval heads in terms of their sensitivity. Sensitivity refers to how a head activates in specific contexts; a highly sensitive head activates in only one context and not in others. For instance, if a passkey context is "the passkey is 1234," a head with high sensitivity would activate strongly for "1" but not for "234." To evaluate this, we compare activation frequency with the average retrieval score, as in previous studies~\cite{wu2024retrieval}. Activation frequency measures how often activation occurs for any token, giving a score of 1 if any token activates. If activation occurs in just one token, the average retrieval score would be 1/4. Therefore, the difference between frequency and average retrieval score indicates whether a head performs roles beyond retrieval in the context. In Figure~\ref{fig:activation_frequency}, we observe a significant difference in the frequency and average score for the retrieval head, suggesting that it plays additional roles outside of retrieval. This is what we refer to as "dynamically activated." On the other hand, the OCR head shows little difference, indicating that these heads, specialized for OCR tasks, are "statically activated."

\vspace{-2mm}
\section{Unveiling the Functional Roles of OCR Heads in Downstream Applications}

To compare the impact of OCR heads and retrieval heads on downstream tasks, we conduct three complementary analyses: CoT prompting, attention head masking, and sink token redistribution. We also use the OCR-VQA~\cite{mishraICDAR19}, Multi-Page DocVQA~\cite{tito2023hierarchical}, DocVQA~\cite{mathew2021docvqa}, NQ~\cite{kwiatkowski-etal-2019-natural}, and HotpotQA~\cite{yang-etal-2018-hotpotqa} datasets for the downstream tasks. Furthermore, the evaluation metric for all the datasets in the downstream tasks is the F1 score.

\begin{table}[t]
\small
\centering
\begin{tabular}{c|ll}
\toprule
\multicolumn{1}{l|}{Qwen2-VL-7B} & \multicolumn{1}{c}{Method} & \multicolumn{1}{c}{Score} \\ \midrule

\multirow{2}{*}{\begin{tabular}[c]{@{}c@{}}Top10\\ OCR Head\end{tabular}}& Avg OCR Score& 0.4004\\
& Avg Retrieval Score& 0.0123\\
\midrule
\multirow{2}{*}{\begin{tabular}[c]{@{}c@{}}Top10\\ Retrieval Head\end{tabular}} & Avg OCR Score& 0.2929\\
& Avg Retrieval Score& 0.4705\\ \bottomrule
\end{tabular}
\vspace{-1mm}
\caption{The average scores of the top 10 OCR and Retrieval heads when CoT prompting is applied in the OCR-VQA task.}
\label{tab:ocr_retrieval_score}
\end{table}

\begin{table*}[ht]
\centering
\small
\resizebox{\linewidth}{!}{\begin{tabular}{l|ccc|ccc|ccc|ccc}
\toprule
Masked Heads& \multicolumn{3}{c|}{DocVQA}& \multicolumn{3}{c|}{MP-DocVQA} & \multicolumn{3}{c|}{NQ} & \multicolumn{3}{c}{HotpotQA}\\ \midrule
\midrule
Qwen2-VL-7B & OCR & RH& Random& OCR & RH& Random& OCR & RH& Random& OCR & RH& Random\\ \midrule
Baseline & \multicolumn{3}{c|}{76.3} & \multicolumn{3}{c|}{69.1} & \multicolumn{3}{c|}{58.0} & \multicolumn{3}{c}{49.6}\\ \midrule
5 & 74.1& 76.1& 75.2 ± 0.3 & 68.1& 69.4& 67.9 ± 0.2 & 57.7& 57.2& 57.6 ± 0.6& 51.2& 47.8& 49.5 ± 0.1\\
10& 73.9& 74.6& 74.2 ± 0.2& 66.7& 67.6& 66.6 ± 0.4& 59.0& 53.8& 56.4 ± 0.2& 49.7& 45.3& 47.9 ± 1.2\\
20& 59.7& 67.1& 75.1 ± 0.4& 47.7& 62.4& 67.7 ± 1.7& 56.8& 36.4& 57.3 ± 0.5& 47.5& 37.3& 48.9 ± 0.4\\ \midrule
\midrule
Intern VL2-8B & OCR & RH& Random& OCR & RH& Random& OCR & RH& Random& OCR & RH& Random\\ \midrule
Baseline & \multicolumn{3}{c|}{39.1} & \multicolumn{3}{c|}{35.0} & \multicolumn{3}{c|}{58.3} & \multicolumn{3}{c}{35.8}\\ \midrule
5 & { 36.6} & { 39.7} & { 39.2 ± 0.2} & { 32.8} & { 35.5} & { 35.1 ± 0.6} & { 60.3} & { 56.3} & { 55.9 ± 0.3} & { 33.5} & { 32.8} & { 34.6 ± 0.1} \\
10& { 30.0} & { 39.7} & { 38.8 ± 0.2} & { 25.1} & { 35.7} & { 34.7 ± 0.4} & { 59.6} & { 55.8} & { 57.8 ± 0.3} & { 33.5} & { 28.2} & { 29.2 ± 0.3} \\
20& { 24.2} & { 38.2} & { 38.3 ± 0.6} & { 20.7} & { 36.1} & { 34.6 ± 0.3} & { 56.6} & { 58.9} & { 57.6 ± 0.2} & { 34.8} & { 25.7} & { 33.5 ± 1.2} \\ \bottomrule
\end{tabular}}
\vspace{-1mm}
\caption{Performance on VQA and QA datasets when masking the OCR head, the retrieval head (RH), or a randomly selected head (Random: mean ± variance across 5 selections).}
\label{tab:masking_7b}
\end{table*}

\subsection{Impact of OCR Head for Reasoning in OCR-VQA}
We investigate the contribution of OCR and retrieval heads to the reasoning capabilities of LVLMs within the OCR-VQA task using CoT prompting. As illustrated in Figure~\ref{fig:downstream}, CoT prompting enables the generation of intermediate reasoning steps before producing the final answer based on the input of the image. The blue and green lines represent the positions where maximum attention occurs in the tokens. We calculate the OCR score between the image tokens and the reasoning part using the method in Section~\ref{method}, and calculate the retrieval score by copying the answer from the reasoning and answer sections, as described in Section~\ref{preliminaries}. In Table~\ref{tab:ocr_retrieval_score}, we measure the scores based on the top 10 heads with the highest average scores for each method. The OCR head shows a high OCR score, while the retrieval head not only shows a high retrieval score but also a high OCR score. The fact that the OCR head only shows a high OCR score, as discussed in Section~\ref{specialized}, suggests that it is more specialized. On the other hand, the retrieval head consistently shows high scores because it is dynamically activated as we discuss in Section~\ref{statically}. This means that the retrieval head is context-sensitive and performs multiple roles. Based on these observations, we highlight the significant influence of the OCR head when performing CoT reasoning on text information in LVLMs.

\subsection{Dissecting the Impact of Attention Head Masking on Task Performance}
To understand the functional contributions of OCR heads and retrieval heads across multiple downstream tasks, we perform a systematic head-masking analysis. Specifically, we evaluate how LVLM performance degrades when we mask the top-scoring OCR heads, retrieval heads (RH), or an equal number of randomly selected heads. We apply this intervention to two VQA tasks—DocVQA, which requires copying specific text from a single image, and Multi-Page DocVQA, which involves reasoning over multiple images—and two QA tasks, Natural Questions (NQ) and HotpotQA, where we include the evidence passage and randomly sample 0–5 negative passages with shuffled ordering. We mask the top 5, 10, and 20 heads of each type (OCR, RH, and random). As shown in Table~\ref{tab:masking_7b}, masking OCR heads causes a pronounced decline in VQA performance: with 20 heads masked, DocVQA and MP-DocVQA F1 scores fall to 59.7\% and 47.7\% (nearly a 20\% drop), whereas masking RH yields much smaller declines of 9.2\% and 7.3\%, respectively. By contrast, on the QA tasks, masking RH leads to a substantial performance decrease. These results confirm that vision–text VQA tasks depend critically on OCR heads, while retrieval heads play a more pivotal role in text-based QA.

\begin{figure*}
    \centering
    \includegraphics[width=0.95\linewidth]{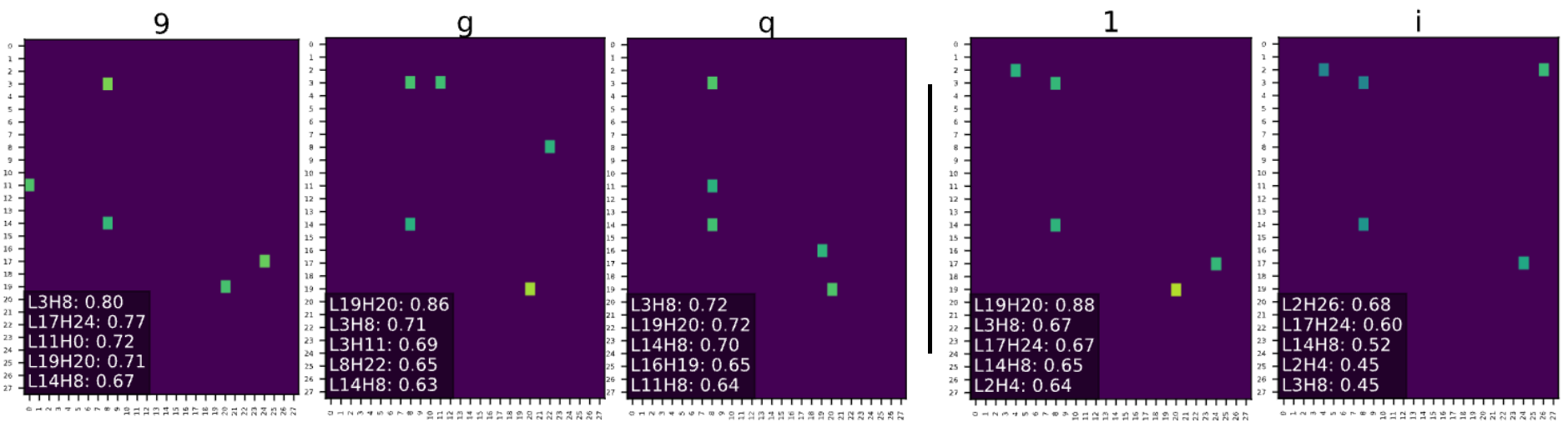}
    \caption{Character‐wise activation of top‐5 OCR heads; visually similar shapes co‐activate common heads.}
    \label{fig:casestudy}
\end{figure*}
\begin{figure*}
    \centering
    \includegraphics[width=0.90\linewidth]{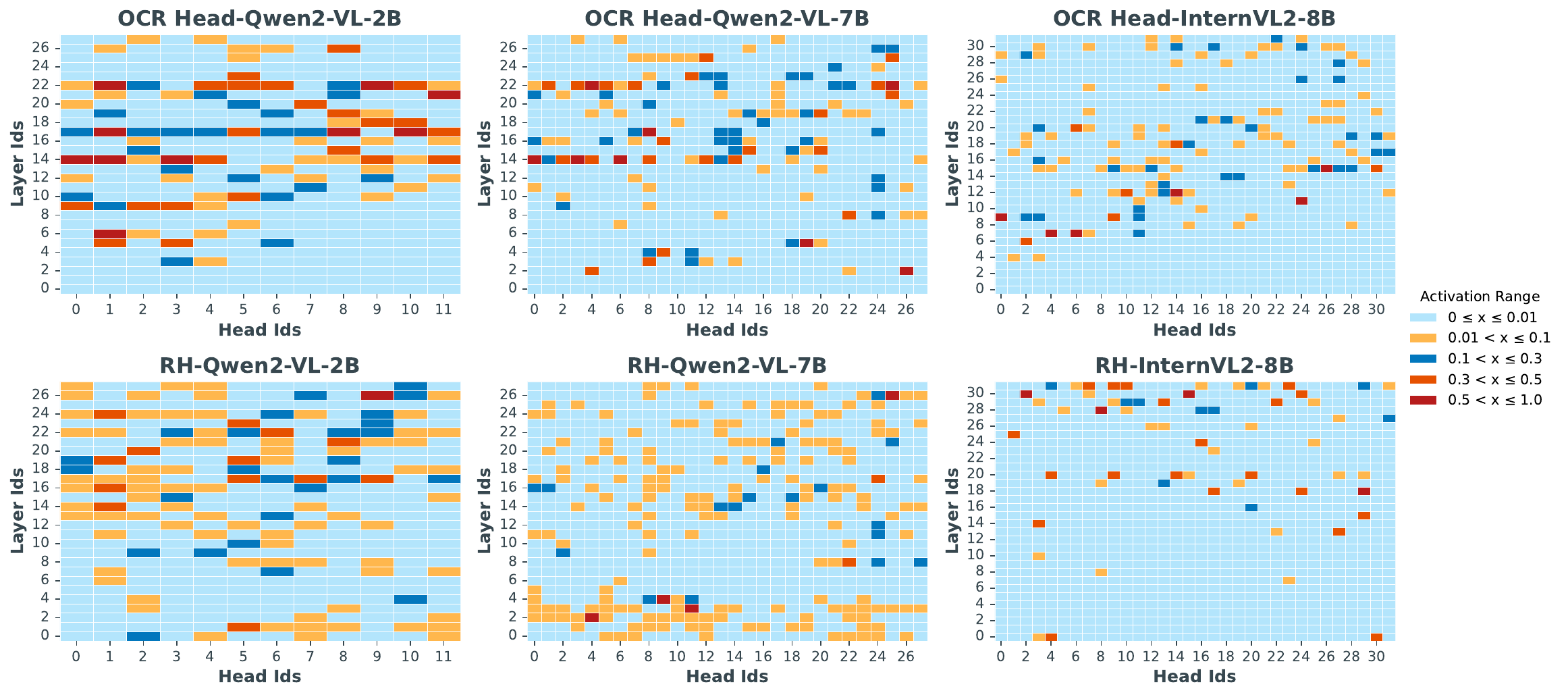}
    \caption{Heatmap visualization of OCR and Retrieval heads.}
    \label{fig:lvlm_heatmap}
\end{figure*}

\begin{table}[t]
\centering
\small
\resizebox{\linewidth}{!}{\begin{tabular}{clcc}
\toprule
\multicolumn{1}{l}{} &\multicolumn{1}{l|}{} & DocVQA & MP-DocVQA \\ \midrule \midrule
\multicolumn{1}{c|}{\multirow{2}{*}{Qwen2-VL-7B}}& \multicolumn{1}{l|}{Baseline} & 76.3 & 69.1\\
\multicolumn{1}{c|}{}& \multicolumn{1}{l|}{OCR} & 76.6 \color{red}{(+0.3)} & 69.7 \color{red}{(+0.6)}\\
\multicolumn{1}{c|}{}& \multicolumn{1}{l|}{RH} & 76.2 \color{blue}{(-0.1)}& 69.4 \color{red}{(+0.3)}\\
\midrule
\multicolumn{1}{c|}{\multirow{2}{*}{InternVL2-8B}} & \multicolumn{1}{l|}{Baseline} & 39.1 & 35.0\\
\multicolumn{1}{c|}{}& \multicolumn{1}{l|}{OCR} & 40.0 \color{red}{(+0.9)} & 35.8 \color{red}{(+0.8)}\\ 
\multicolumn{1}{c|}{}& \multicolumn{1}{l|}{RH} & 39.3 \color{red}{(+0.2)}&  34.7 \color{blue}{(-0.3)}\\ 
\bottomrule
\end{tabular}}
\caption{Performance results after redistributing the attention sink token value in the OCR head and retrieval head (RH) across MP-DocVQA and DocVQA tasks.}

\label{tab:redistribution}

\end{table}

\subsection{Enhancing OCR Head Performance via Sink Token Value Redistribution}
In this experiment, we investigate the potential for enhancing downstream task performance by strategically redistributing the attention values associated with the sink token.
An attention sink token~\cite{kang2025see} refers to a token in the input sequence that receives disproportionately high attention scores despite its lack of semantic relevance. Although it carries little meaningful information, its attention value diverts focus from more important tokens in the sequence. Recent research~\cite{10.5555/3692070.3694448, kang2025see, gu2025when} suggests that optimizing the attention score distribution and reducing the impact of such attention sinks can enhance performance, especially in tasks requiring more accurate and meaningful attention allocation. Based on this, we redistribute the sink token value proportionally to the attention scores of other tokens. This process can be expressed as:

\begin{equation}
\resizebox{0.42\textwidth}{!}{
$\hat{A}^l_h[T, t] = A^l_h [T, t] + \beta \cdot S \cdot \frac{ A^l_h[T, t]}{\sum_{i \in \{1, \cdots, T\} } A^l_h[T, i] }
$}
\label{eq:attention}
\end{equation}
Here, $l$ and $h$ refer to the layer and head, respectively, and $A$ represents the attention score. $T$ refers to the total number of tokens in the attention score, and $S$ is the original value of the sink token. $\beta$ is a hyperparameter that controls how much we want to remove excess attention scores from the attention sink. Following prior studies~\cite{10.5555/3692070.3694448}, we set $\beta$ = 0.4. First, we compute the attention score ratios for all tokens, excluding the 0th sink token. The computed ratios are then multiplied by $S$ to calculate the redistributed attention scores for each token, which are added to the original attention score to obtain $\hat{A}$. This new $\hat{A}$ is used for subsequent operations, and after inference, the F1 score is measured. The heads selected for redistribution are the top 4 heads with the highest average OCR or retrieval scores. Table~\ref{tab:redistribution} shows that redistributing the sink token among OCR heads boosts performance on DocVQA and MP-DocVQA for both QwenVL and InternVL models. In contrast, redistributing among retrieval heads lowers performance in some cases and yields slight gains in others.

\section{Case Study}  
\subsection{Explaining Character Recognition via OCR Heads}
We investigate whether each character—from 0 to 9 and a to z—tends to activate a specific attention head. To do this, we replaced the original multi-digit pass keys in our dataset with single characters (0–9, a–z), rendered the prompts as images, and then applied the OCR scoring procedure described in Section~\ref{method}. Figure~\ref{fig:casestudy} shows the activation heatmaps along with the top‐5 heads per character. Characters with similar shapes consistently activate the same OCR heads—for instance, ‘1’ and ‘i’ share four heads (L3H8, L2H4, L14H8, and L17H24), and ‘9’, ‘g’, and ‘q’ share three. This consistent co‐activation demonstrates that OCR heads mechanistically encode visual similarity to guide text extraction. We show that the behavior of OCR heads can directly elucidate the mechanism by which LVLMs perform optical character recognition on embedded text. The complete set of character‐wise activation heatmaps for all tokens is presented in Appendix~\ref{appen:char}.

\subsection{Distribution of OCR and Retrieval Heads}

In Figure~\ref{fig:lvlm_heatmap}, we visualize the distribution of OCR and Retrieval heads. The OCR heads are not uniformly distributed; rather, they are found to be concentrated in specific areas of the middle-to-late layers. Figure~\ref{fig:lvlm_heatmap} displays the number of OCR heads per layer (average score > 0.1) for the Qwen2-VL series, revealing a clear concentration. Specifically, in the Qwen2-VL-2B model, there are 11 OCR heads in the 17th layer and 8 OCR heads in the 22nd layer. Similarly, for the Qwen2-VL-7B model, 10 OCR heads are located in the 14th layer and 11 OCR heads in the 22nd layer. This demonstrates that OCR heads tend to cluster within certain middle-to-late layers, rather than being evenly distributed across all layers.
\section{Related Work}

\paragraph{Attention Head Analysis in LVLM}
Interpretability research in LVLMs has investigated how attention mechanisms integrate visual and textual information. These studies have identified different types of heads involved in processing visual input: Image-centric heads~\cite{kang2025see} focus on semantically meaningful visual features and manage attention distribution. The other work~\cite{liu2024paying} explores amplifying visual attention to reduce issues like hallucination. Localization heads~\cite{kang2025your} are shown to be effective for visual grounding, pinpointing image regions based on text queries without explicit training. Furthermore, Visual heads actively processing visual information have been identified using entropy-based scores~\cite{bi2024unveiling}. While these studies shed light on LVLM's visual processing and visual-text alignment, they do not specifically investigate attention heads dedicated to the unique task of recognizing and retrieving text embedded within images. Understanding these multimodal mechanisms builds upon interpretability insights gained from analyzing LLMs.

\paragraph{In-Context Identification Heads}
Transformer-based models utilize attention heads that focus on different aspects of input data to extract meaningful information~\cite{zheng2024attention}. In‑Context Identification heads are attention heads that selectively attend to and aggregate structural, syntactic, and semantic cues within the prompt, writing these distilled signals back into the residual stream to guide subsequent generation. For example, the Previous Head~\cite{olsson2022context} captures positional relationships within token sequences, while the Duplicate Head~\cite{wang2022interpretability} identifies repeating content, becoming more active with increased repetition of tokens. In contrast, the Rare Words Head~\cite{voita2019analyzing} targets tokens with low frequency. Additionally, Retrieval Heads~\cite{wu2024retrieval} precisely pinpoint the positions of specific tokens. Identifying these specialized attention heads typically involves replacement-based methods such as ablation~\cite{yu2024large, jin2024cutting} or patching~\cite{merullo2023circuit, todd2023function}, which test their influence on model outputs, or scoring-based methods~\cite{jin2024cutting, crosbie2024induction} that evaluate their functionality. In this study, we propose the OCR Score, a scoring-based approach for efficiently detecting attention heads with unique capabilities.

\section{Conclusion}


This paper identifies and characterizes Optical Character Recognition (OCR) heads as the specialized mechanisms that Large Vision-Language Models (LVLMs) use to read text directly from images. We demonstrate that these heads are functionally distinct from conventional retrieval heads, defined by three key properties: lower sparsity, qualitative distinctiveness, and static activation patterns. Our experiments validate their critical role; targeted head-masking severely degrades performance on OCR-VQA tasks while leaving text-based QA unaffected, confirming their specialized function. Furthermore, we show that manipulating these heads via sink-token redistribution can enhance performance on downstream tasks. This work provides a direct, mechanistic insight into how LVLMs process embedded text, paving the way for more interpretable and reliable multimodal models.

\section*{Limitations}
We validate the OCR head properties through five empirical experiments. However, many open-source LVLMs support only a single image input or rely on thumbnail-style training settings. Some models also produce image tokens that mismatch with patch counts due to encoder variations. These constraints limit the diversity of models we evaluate. Additionally, due to computational constraints, we conduct experiments on the passkey and NIAH dataset only up to 8k samples.
\section*{Ethics Statement}
This study involves open-source LVLMs for QA tasks, and some models may generate inappropriate outputs. Also, the VQA and QA datasets used in our experiments may also include potentially harmful or biased content.

\section*{Acknowledgments}
This work was supported by the Institute of Information \& Communications Technology Planning \& Evaluation (IITP) grant funded by the Korea government (MSIT) [RS-2021-II211341, Artificial Intelligence Graduate School Program (Chung-Ang University)] and Korea Institute for Advancement of Technology(KIAT) grant funded by the Korea Government(MOTIE) (RS-2025-25458133).

\bibliography{custom}

\begin{thebibliography}{39}
\providecommand{\natexlab}[1]{#1}

\bibitem[{Achiam et~al.(2023)Achiam, Adler, Agarwal, Ahmad, Akkaya, Aleman,
  Almeida, Altenschmidt, Altman, Anadkat et~al.}]{achiam2023gpt}
Josh Achiam, Steven Adler, Sandhini Agarwal, Lama Ahmad, Ilge Akkaya,
  Florencia~Leoni Aleman, Diogo Almeida, Janko Altenschmidt, Sam Altman,
  Shyamal Anadkat, et~al. 2023.
\newblock Gpt-4 technical report.
\newblock \emph{arXiv preprint arXiv:2303.08774}.

\bibitem[{Antol et~al.(2015)Antol, Agrawal, Lu, Mitchell, Batra, Zitnick, and
  Parikh}]{antol2015vqa}
Stanislaw Antol, Aishwarya Agrawal, Jiasen Lu, Margaret Mitchell, Dhruv Batra,
  C~Lawrence Zitnick, and Devi Parikh. 2015.
\newblock Vqa: Visual question answering.
\newblock In \emph{Proceedings of the IEEE international conference on computer
  vision}, pages 2425--2433.

\bibitem[{Bi et~al.(2024)Bi, Guo, Tang, Wen, Liu, and Xu}]{bi2024unveiling}
Jing Bi, Junjia Guo, Yunlong Tang, Lianggong~Bruce Wen, Zhang Liu, and
  Chenliang Xu. 2024.
\newblock Unveiling visual perception in language models: An attention head
  analysis approach.
\newblock \emph{arXiv preprint arXiv:2412.18108}.

\bibitem[{Cai et~al.(2024)Cai, Cao, Chen, Chen, Chen, Chen, Chen, Chen, Chen,
  Chu et~al.}]{cai2024internlm2}
Zheng Cai, Maosong Cao, Haojiong Chen, Kai Chen, Keyu Chen, Xin Chen, Xun Chen,
  Zehui Chen, Zhi Chen, Pei Chu, et~al. 2024.
\newblock Internlm2 technical report.
\newblock \emph{arXiv preprint arXiv:2403.17297}.

\bibitem[{Crosbie and Shutova(2024)}]{crosbie2024induction}
Joy Crosbie and Ekaterina Shutova. 2024.
\newblock Induction heads as an essential mechanism for pattern matching in
  in-context learning.
\newblock \emph{arXiv preprint arXiv:2407.07011}.

\bibitem[{Dong et~al.(2024)Dong, Zhang, Zang, Cao, Wang, Ouyang, Wei, Zhang,
  Duan, Cao, Zhang, Li, Yan, Gao, Zhang, Li, Li, Chen, He, Zhang, Qiao, Lin,
  and Wang}]{DBLP:journals/corr/abs-2401-16420}
Xiaoyi Dong, Pan Zhang, Yuhang Zang, Yuhang Cao, Bin Wang, Linke Ouyang, Xilin
  Wei, Songyang Zhang, Haodong Duan, Maosong Cao, Wenwei Zhang, Yining Li, Hang
  Yan, Yang Gao, Xinyue Zhang, Wei Li, Jingwen Li, Kai Chen, Conghui He,
  Xingcheng Zhang, Yu~Qiao, Dahua Lin, and Jiaqi Wang. 2024.
\newblock \href {https://doi.org/10.48550/arXiv.2401.16420}
  {Internlm-xcomposer2: Mastering free-form text-image composition and
  comprehension in vision-language large model}.
\newblock \emph{CoRR}, abs/2401.16420.

\bibitem[{Goyal et~al.(2017)Goyal, Khot, Summers-Stay, Batra, and
  Parikh}]{goyal2017making}
Yash Goyal, Tejas Khot, Douglas Summers-Stay, Dhruv Batra, and Devi Parikh.
  2017.
\newblock Making the v in vqa matter: Elevating the role of image understanding
  in visual question answering.
\newblock In \emph{Proceedings of the IEEE conference on computer vision and
  pattern recognition}, pages 6904--6913.

\bibitem[{Grattafiori et~al.(2024)Grattafiori, Dubey, Jauhri, Pandey, Kadian,
  Al-Dahle, Letman, Mathur, Schelten, Vaughan et~al.}]{grattafiori2024llama}
Aaron Grattafiori, Abhimanyu Dubey, Abhinav Jauhri, Abhinav Pandey, Abhishek
  Kadian, Ahmad Al-Dahle, Aiesha Letman, Akhil Mathur, Alan Schelten, Alex
  Vaughan, et~al. 2024.
\newblock The llama 3 herd of models.
\newblock \emph{arXiv preprint arXiv:2407.21783}.

\bibitem[{Gu et~al.(2025)Gu, Pang, Du, Liu, Zhang, Du, Wang, and
  Lin}]{gu2025when}
Xiangming Gu, Tianyu Pang, Chao Du, Qian Liu, Fengzhuo Zhang, Cunxiao Du,
  Ye~Wang, and Min Lin. 2025.
\newblock \href {https://openreview.net/forum?id=78Nn4QJTEN} {When attention
  sink emerges in language models: An empirical view}.
\newblock In \emph{The Thirteenth International Conference on Learning
  Representations}.

\bibitem[{Huang et~al.(2024)Huang, Chan, Fung, Qiu, Zhou, Joty, Chang, and
  Ji}]{huang2024pixels}
Kung-Hsiang Huang, Hou~Pong Chan, Yi~R Fung, Haoyi Qiu, Mingyang Zhou, Shafiq
  Joty, Shih-Fu Chang, and Heng Ji. 2024.
\newblock From pixels to insights: A survey on automatic chart understanding in
  the era of large foundation models.
\newblock \emph{IEEE Transactions on Knowledge and Data Engineering}.

\bibitem[{Jin et~al.(2024)Jin, Cao, Yuan, Chen, Xu, Li, Jiang, Liu, and
  Zhao}]{jin2024cutting}
Zhuoran Jin, Pengfei Cao, Hongbang Yuan, Yubo Chen, Jiexin Xu, Huaijun Li,
  Xiaojian Jiang, Kang Liu, and Jun Zhao. 2024.
\newblock Cutting off the head ends the conflict: A mechanism for interpreting
  and mitigating knowledge conflicts in language models.
\newblock \emph{arXiv preprint arXiv:2402.18154}.

\bibitem[{Kamradt(2023)}]{needle-in-haystack}
Gregory Kamradt. 2023.
\newblock \href {https://github.com/gkamradt/LLMTest_NeedleInAHaystack} {Needle
  in a haystack - pressure testing llms}.

\bibitem[{Kang et~al.(2025{\natexlab{a}})Kang, Kim, Kim, and
  Hwang}]{kang2025see}
Seil Kang, Jinyeong Kim, Junhyeok Kim, and Seong~Jae Hwang. 2025{\natexlab{a}}.
\newblock \href {https://openreview.net/forum?id=7uDI7w5RQA} {See what you are
  told: Visual attention sink in large multimodal models}.
\newblock In \emph{The Thirteenth International Conference on Learning
  Representations}.

\bibitem[{Kang et~al.(2025{\natexlab{b}})Kang, Kim, Kim, and
  Hwang}]{kang2025your}
Seil Kang, Jinyeong Kim, Junhyeok Kim, and Seong~Jae Hwang. 2025{\natexlab{b}}.
\newblock Your large vision-language model only needs a few attention heads for
  visual grounding.
\newblock \emph{arXiv preprint arXiv:2503.06287}.

\bibitem[{Kwiatkowski et~al.(2019)Kwiatkowski, Palomaki, Redfield, Collins,
  Parikh, Alberti, Epstein, Polosukhin, Devlin, Lee, Toutanova, Jones, Kelcey,
  Chang, Dai, Uszkoreit, Le, and Petrov}]{kwiatkowski-etal-2019-natural}
Tom Kwiatkowski, Jennimaria Palomaki, Olivia Redfield, Michael Collins, Ankur
  Parikh, Chris Alberti, Danielle Epstein, Illia Polosukhin, Jacob Devlin,
  Kenton Lee, Kristina Toutanova, Llion Jones, Matthew Kelcey, Ming-Wei Chang,
  Andrew~M. Dai, Jakob Uszkoreit, Quoc Le, and Slav Petrov. 2019.
\newblock \href {https://doi.org/10.1162/tacl_a_00276} {Natural questions: A
  benchmark for question answering research}.
\newblock \emph{Transactions of the Association for Computational Linguistics},
  7:452--466.

\bibitem[{Li et~al.(2023)Li, Li, Savarese, and Hoi}]{pmlr-v202-li23q}
Junnan Li, Dongxu Li, Silvio Savarese, and Steven Hoi. 2023.
\newblock \href {https://proceedings.mlr.press/v202/li23q.html} {{BLIP}-2:
  Bootstrapping language-image pre-training with frozen image encoders and
  large language models}.
\newblock In \emph{Proceedings of the 40th International Conference on Machine
  Learning}, volume 202 of \emph{Proceedings of Machine Learning Research},
  pages 19730--19742. PMLR.

\bibitem[{Li et~al.(2025)Li, Liu, Li, Zhang, Xu, Chen, Shi, Jiang, Wang, Wang
  et~al.}]{li2025perception}
Yunxin Li, Zhenyu Liu, Zitao Li, Xuanyu Zhang, Zhenran Xu, Xinyu Chen, Haoyuan
  Shi, Shenyuan Jiang, Xintong Wang, Jifang Wang, et~al. 2025.
\newblock Perception, reason, think, and plan: A survey on large multimodal
  reasoning models.
\newblock \emph{arXiv preprint arXiv:2505.04921}.

\bibitem[{Liu et~al.(2024{\natexlab{a}})Liu, Li, Li, and Lee}]{Liu_2024_CVPR}
Haotian Liu, Chunyuan Li, Yuheng Li, and Yong~Jae Lee. 2024{\natexlab{a}}.
\newblock Improved baselines with visual instruction tuning.
\newblock In \emph{Proceedings of the IEEE/CVF Conference on Computer Vision
  and Pattern Recognition (CVPR)}, pages 26296--26306.

\bibitem[{Liu et~al.(2023)Liu, Li, Wu, and Lee}]{liu2023llava}
Haotian Liu, Chunyuan Li, Qingyang Wu, and Yong~Jae Lee. 2023.
\newblock Visual instruction tuning.
\newblock In \emph{NeurIPS}.

\bibitem[{Liu et~al.(2024{\natexlab{b}})Liu, Zheng, and Chen}]{liu2024paying}
Shi Liu, Kecheng Zheng, and Wei Chen. 2024{\natexlab{b}}.
\newblock Paying more attention to image: A training-free method for
  alleviating hallucination in lvlms.
\newblock In \emph{European Conference on Computer Vision}, pages 125--140.
  Springer.

\bibitem[{Mathew et~al.(2021)Mathew, Karatzas, and Jawahar}]{mathew2021docvqa}
Minesh Mathew, Dimosthenis Karatzas, and CV~Jawahar. 2021.
\newblock Docvqa: A dataset for vqa on document images.
\newblock In \emph{Proceedings of the IEEE/CVF winter conference on
  applications of computer vision}, pages 2200--2209.

\bibitem[{Merullo et~al.(2023)Merullo, Eickhoff, and
  Pavlick}]{merullo2023circuit}
Jack Merullo, Carsten Eickhoff, and Ellie Pavlick. 2023.
\newblock Circuit component reuse across tasks in transformer language models.
\newblock \emph{arXiv preprint arXiv:2310.08744}.

\bibitem[{Mishra et~al.(2019)Mishra, Shekhar, Singh, and
  Chakraborty}]{mishraICDAR19}
Anand Mishra, Shashank Shekhar, Ajeet~Kumar Singh, and Anirban Chakraborty.
  2019.
\newblock Ocr-vqa: Visual question answering by reading text in images.
\newblock In \emph{ICDAR}.

\bibitem[{Olsson et~al.(2022)Olsson, Elhage, Nanda, Joseph, DasSarma, Henighan,
  Mann, Askell, Bai, Chen et~al.}]{olsson2022context}
Catherine Olsson, Nelson Elhage, Neel Nanda, Nicholas Joseph, Nova DasSarma,
  Tom Henighan, Ben Mann, Amanda Askell, Yuntao Bai, Anna Chen, et~al. 2022.
\newblock In-context learning and induction heads.
\newblock \emph{arXiv preprint arXiv:2209.11895}.

\bibitem[{Reddy et~al.(2024)Reddy, Koncel-Kedziorski, Lai, Krumdick, Lovering,
  and Tanner}]{reddy2024docfinqa}
Varshini Reddy, Rik Koncel-Kedziorski, Viet~Dac Lai, Michael Krumdick, Charles
  Lovering, and Chris Tanner. 2024.
\newblock Docfinqa: A long-context financial reasoning dataset.
\newblock \emph{arXiv preprint arXiv:2401.06915}.

\bibitem[{Tito et~al.(2023)Tito, Karatzas, and Valveny}]{tito2023hierarchical}
Rub{\`e}n Tito, Dimosthenis Karatzas, and Ernest Valveny. 2023.
\newblock Hierarchical multimodal transformers for multipage docvqa.
\newblock \emph{Pattern Recognition}, 144:109834.

\bibitem[{Todd et~al.(2023)Todd, Li, Sharma, Mueller, Wallace, and
  Bau}]{todd2023function}
Eric Todd, Millicent~L Li, Arnab~Sen Sharma, Aaron Mueller, Byron~C Wallace,
  and David Bau. 2023.
\newblock Function vectors in large language models.
\newblock \emph{arXiv preprint arXiv:2310.15213}.

\bibitem[{Voita et~al.(2019)Voita, Talbot, Moiseev, Sennrich, and
  Titov}]{voita2019analyzing}
Elena Voita, David Talbot, Fedor Moiseev, Rico Sennrich, and Ivan Titov. 2019.
\newblock Analyzing multi-head self-attention: Specialized heads do the heavy
  lifting, the rest can be pruned.
\newblock \emph{arXiv preprint arXiv:1905.09418}.

\bibitem[{Wang et~al.(2022)Wang, Variengien, Conmy, Shlegeris, and
  Steinhardt}]{wang2022interpretability}
Kevin Wang, Alexandre Variengien, Arthur Conmy, Buck Shlegeris, and Jacob
  Steinhardt. 2022.
\newblock Interpretability in the wild: a circuit for indirect object
  identification in gpt-2 small.
\newblock \emph{arXiv preprint arXiv:2211.00593}.

\bibitem[{Wang et~al.(2024)Wang, Bai, Tan, Wang, Fan, Bai, Chen, Liu, Wang, Ge
  et~al.}]{wang2024qwen2}
Peng Wang, Shuai Bai, Sinan Tan, Shijie Wang, Zhihao Fan, Jinze Bai, Keqin
  Chen, Xuejing Liu, Jialin Wang, Wenbin Ge, et~al. 2024.
\newblock Qwen2-vl: Enhancing vision-language model's perception of the world
  at any resolution.
\newblock \emph{arXiv preprint arXiv:2409.12191}.

\bibitem[{Wei et~al.(2022)Wei, Wang, Schuurmans, Bosma, Xia, Chi, Le, Zhou
  et~al.}]{wei2022chain}
Jason Wei, Xuezhi Wang, Dale Schuurmans, Maarten Bosma, Fei Xia, Ed~Chi, Quoc~V
  Le, Denny Zhou, et~al. 2022.
\newblock Chain-of-thought prompting elicits reasoning in large language
  models.
\newblock \emph{Advances in neural information processing systems},
  35:24824--24837.

\bibitem[{Wu et~al.(2024{\natexlab{a}})Wu, Zhang, Xia, Li, Xia, Chang, Yu, Kim,
  Rossi, Zhang et~al.}]{wu2024visual}
Junda Wu, Zhehao Zhang, Yu~Xia, Xintong Li, Zhaoyang Xia, Aaron Chang, Tong Yu,
  Sungchul Kim, Ryan~A Rossi, Ruiyi Zhang, et~al. 2024{\natexlab{a}}.
\newblock Visual prompting in multimodal large language models: A survey.
\newblock \emph{arXiv preprint arXiv:2409.15310}.

\bibitem[{Wu et~al.(2024{\natexlab{b}})Wu, Wang, Xiao, Peng, and
  Fu}]{wu2024retrieval}
Wenhao Wu, Yizhong Wang, Guangxuan Xiao, Hao Peng, and Yao Fu.
  2024{\natexlab{b}}.
\newblock Retrieval head mechanistically explains long-context factuality.
\newblock \emph{arXiv preprint arXiv:2404.15574}.

\bibitem[{Xiao et~al.(2024)Xiao, Yang, Lan, Wang, and Xu}]{xiao2024towards}
Linhui Xiao, Xiaoshan Yang, Xiangyuan Lan, Yaowei Wang, and Changsheng Xu.
  2024.
\newblock Towards visual grounding: A survey.
\newblock \emph{arXiv preprint arXiv:2412.20206}.

\bibitem[{Yang et~al.(2018)Yang, Qi, Zhang, Bengio, Cohen, Salakhutdinov, and
  Manning}]{yang-etal-2018-hotpotqa}
Zhilin Yang, Peng Qi, Saizheng Zhang, Yoshua Bengio, William Cohen, Ruslan
  Salakhutdinov, and Christopher~D. Manning. 2018.
\newblock \href {https://doi.org/10.18653/v1/D18-1259} {{H}otpot{QA}: A dataset
  for diverse, explainable multi-hop question answering}.
\newblock In \emph{Proceedings of the 2018 Conference on Empirical Methods in
  Natural Language Processing}, pages 2369--2380, Brussels, Belgium.
  Association for Computational Linguistics.

\bibitem[{Yu and Ananiadou(2024)}]{yu2024large}
Zeping Yu and Sophia Ananiadou. 2024.
\newblock How do large language models learn in-context? query and key matrices
  of in-context heads are two towers for metric learning.
\newblock \emph{arXiv preprint arXiv:2402.02872}.

\bibitem[{Yu et~al.(2024)Yu, Wang, Fu, Shi, Shaikh, and
  Lin}]{10.5555/3692070.3694448}
Zhongzhi Yu, Zheng Wang, Yonggan Fu, Huihong Shi, Khalid Shaikh, and
  Yingyan~(Celine) Lin. 2024.
\newblock Unveiling and harnessing hidden attention sinks: enhancing large
  language models without training through attention calibration.
\newblock In \emph{Proceedings of the 41st International Conference on Machine
  Learning}, ICML'24. JMLR.org.

\bibitem[{Zhao et~al.(2023)Zhao, Zhou, Li, Tang, Wang, Hou, Min, Zhang, Zhang,
  Dong et~al.}]{zhao2023survey}
Wayne~Xin Zhao, Kun Zhou, Junyi Li, Tianyi Tang, Xiaolei Wang, Yupeng Hou,
  Yingqian Min, Beichen Zhang, Junjie Zhang, Zican Dong, et~al. 2023.
\newblock A survey of large language models.
\newblock \emph{arXiv preprint arXiv:2303.18223}, 1(2).

\bibitem[{Zheng et~al.(2024)Zheng, Wang, Huang, Song, Yang, Tang, Xiong, and
  Li}]{zheng2024attention}
Zifan Zheng, Yezhaohui Wang, Yuxin Huang, Shichao Song, Mingchuan Yang,
  Bo~Tang, Feiyu Xiong, and Zhiyu Li. 2024.
\newblock Attention heads of large language models: A survey.
\newblock \emph{arXiv preprint arXiv:2409.03752}.

\end{thebibliography}
\appendix
\newpage
\section{Character-wise Analysis}
\label{appen:char}

\begin{figure*}
    \centering
    \includegraphics[width=1.0\linewidth]{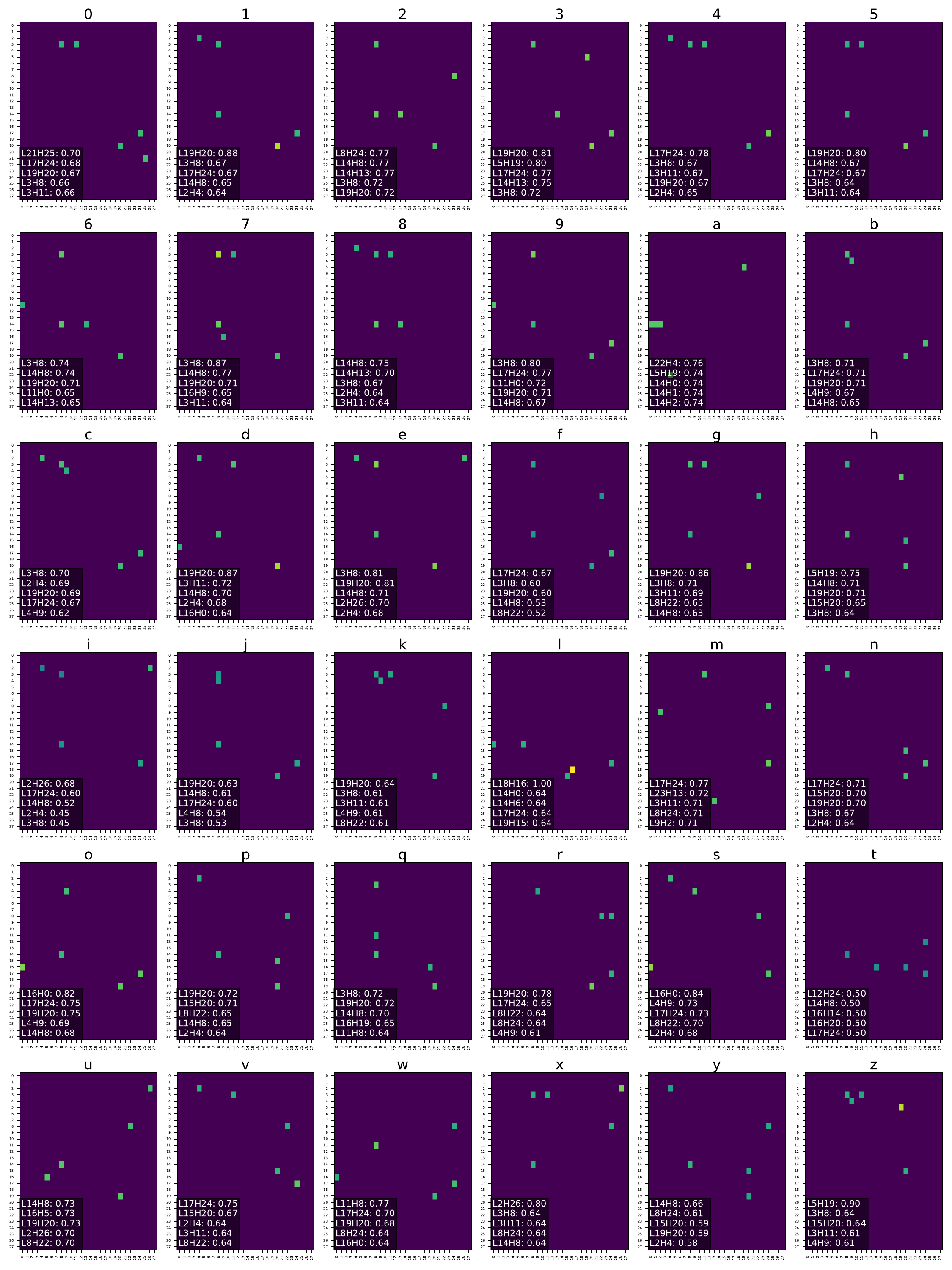}
    \caption{Heatmap visualization of top5 OCR scores for each character from 0 to 9 and a to z.}
    \label{fig:char_heatmap}
\end{figure*}

We investigate whether each character—from 0 to 9 and a to z—tends to activate a specific attention head. To do this, we replaced the original multi-digit pass keys in our dataset with single characters (0–9, a–z), rendered the prompts as images, and then applied the OCR scoring procedure described above. As shown in Figure~\ref{fig:char_heatmap}, we found that visually similar characters share common high-activation heads among their top 5. For instance, for the characters “1” and “i,” four heads (L14H8, L2H4, L17H24, and L3H8) appear jointly in their top 5. Similarly, for “9,” “g,” and “q,” the heads L3H8, L14H8, and L19H20 co-occur among the top 5 activations. This suggests that when visually similar shapes enter the ViT encoder, they produce similar representations that in turn trigger the same decoder heads in the LLM. In Figure 6, we use the Qwen2-VL-7B model; the horizontal axis denotes the head index, and the vertical axis denotes the layer index.  

\section{CoT Prompt}
\begin{table}[!h]
\centering
\scalebox{0.85}{\begin{tabular}{l}
\hline
\begin{tabular}[c]{@{}p{1.0\linewidth}@{}}\\
\textbf{Let's think about it step by step. The response must be a valid JSON object with two fields:} "thinking" and "answer".\\
\\
The "answer" field should provide the final answer, starting with "Answer:".\\
\\
Here is the question:\\
"Question": {questions[n]}\\

Respond in the following JSON format:\\
\{
\\
"\textbf{thinking}": "Thinking: ...",\\
"\textbf{answer}": "Answer: ..."\\
\}
\\
\end{tabular}  \\ \hline
\end{tabular}}
\caption{CoT Prompt.}
\label{tab:prompt}
\end{table}

In Section~\ref{statically}, we use Chain-of-Thought (CoT) prompting to simultaneously measure the OCR score and the retrieval score. To obtain structured responses, we instruct the model to generate reasoning and answers in JSON format, as illustrated in Table~\ref{tab:prompt}. Specifically, we extract the OCR score from the input image tokens and reasoning component, and compute the retrieval score from the reasoning and answer components.

\end{document}